\documentclass{article}

\usepackage{PRIMEarxiv}
\usepackage{amsmath}
\usepackage[utf8]{inputenc} 
\usepackage[T1]{fontenc}    
\usepackage{hyperref}       
\usepackage{url}            
\usepackage{booktabs}       
\usepackage{amsfonts}       
\usepackage{nicefrac}       
\usepackage{microtype}      
\usepackage{lipsum}
\usepackage{fancyhdr}       
\usepackage{graphicx}       
\graphicspath{{media/}}     

\title{Scalable Whole Slide Image Representation Using K-means Clustering and Fisher Vector Aggregation
}

\author{
Ravi Kant Gupta \hspace{5mm} Shounak Das \hspace{5mm} Ardhendu Sekhar \hspace{5mm}
\textbf{Amit Sethi}\\
  Department of Electrical Engineering,
  Indian Institute of Technology Bombay \\
  \texttt{\{ravigupta131, 21D070068, asekhar, asethi\}@iitb.ac.in} \\
}

\begin{document}
\maketitle

\begin{abstract}
Whole slide images (WSIs) are high-resolution, gigapixel-sized images that pose significant computational challenges for traditional machine learning models due to their size and heterogeneity. In this paper, we present a scalable and efficient methodology for WSI classification by leveraging patch-based feature extraction, clustering, and Fisher vector encoding. Initially, WSIs are divided into fixed-size patches, and deep feature embeddings are extracted from each patch using a pre-trained convolutional neural network (CNN). These patch-level embeddings are subsequently clustered using K-means clustering, where each cluster aggregates semantically similar regions of the WSI. To effectively summarize each cluster, Fisher vector representations are computed by modeling the distribution of patch embeddings in each cluster as a parametric Gaussian mixture model (GMM). The Fisher vectors from each cluster are concatenated into a high-dimensional feature vector, creating a compact and informative representation of the entire WSI. This feature vector is then used by a classifier to predict the WSI’s diagnostic label. Our method captures local and global tissue structures and yields robust performance for large-scale WSI classification, demonstrating superior accuracy and scalability compared to other approaches.
\end{abstract}


\section{Introduction}
Whole slide images (WSIs) are used in digital pathology for diagnosis and research, offering detailed histological views of entire tissue sections. Despite their value, their large size and complexity pose significant challenges for machine learning and deep learning, particularly in terms of computational and memory requirements. Efficient methods to process and classify WSIs are critical for advancing automated pathology. A common approach to handling WSIs is to divide them into smaller patches for independent processing, reducing computational load. However, this can overlook broader structural information. Recent research addresses this by extracting and aggregating patch-level features to capture a more comprehensive representation of the entire WSI.

In this study, we present a framework for WSI classification that combines patch-based feature extraction, clustering, and Fisher vector encoding to create a compact yet detailed representation of the entire WSI. The WSI is divided into fixed-size patches, and feature embeddings are extracted using pre-trained CNNs~\cite{albawi2017understanding} and transformers~\cite{khan2022transformers}. K-means clustering~\cite{lloyd1982least} is used to group similar patches, and Fisher vectors~\cite{akbarnejad2021deep} are computed for each cluster using a GMM~\cite{Arun2020}, which are then concatenated into a high-dimensional feature vector capturing local and global information.

This vector is fed into a classifier to predict the WSI’s diagnostic label, effectively combining patch-level features with global context for large-scale WSI classification. We validated our method on multiple datasets: Warwick~\cite{warwicket} for HER2 scoring and HER2+ vs HER2- classification, TCGA-BRCA~\cite{tcga_brca} for HER2 classification, TCGA-LUAD for mutation prediction (EGFR vs Non-EGFR)~\cite{gupta2023egfr}, and CAMELYON17 dataset~\cite{Bandi2018et} for metastasis detection. Across all datasets, the proposed approach consistently outperformed traditional patch-based methods while maintaining computational efficiency across diverse classification tasks and medical imaging modalities.

\section{Methodology}

In this study, we propose a method for classifying whole slide images (WSIs) by extracting features from patches, clustering these features, and encoding them using Fisher vectors. The following steps outline the entire process:

\begin{figure*}[!]
\centering
\includegraphics[height=6.5cm,width=12cm]{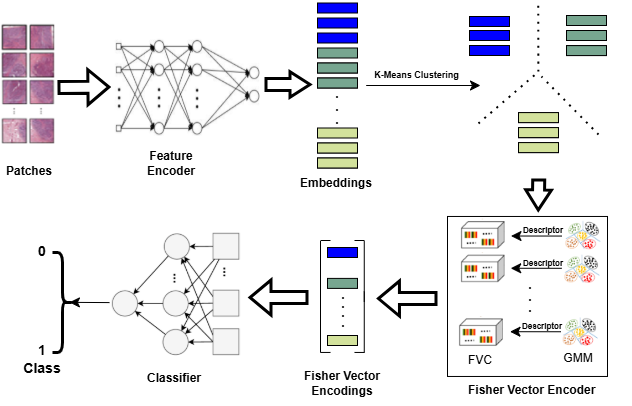}
\vspace{0.25cm}
\caption{Patches from WSIs are encoded into embeddings, clustered via K-means, and transformed into Fisher vectors using a GMM model. These Fisher vectors are concatenated into a single vector representing the WSI, which is then used for classification.}
\label{fig1}
\end{figure*}

We begin by preprocessing WSIs using ~\cite{patil2023efficient} model to remove artifacts and non-tissue regions, ensuring that only tissue areas are retained for further analysis. After getting quality tissue regions, we extract 512×512 pixel patches.

For each extracted patch, we compute feature embeddings using pre-trained networks. We experimented with a variety of deep learning models, including convolutional neural networks (CNNs) and transformer-based architectures. 

While patch-level feature extractors capture local features, our key ideas for complementing those with global features are the following. After obtaining patch-level feature embeddings, firstly, we apply K-means clustering to group patches based on feature similarity, thus reducing dataset complexity. The optimal number of clusters is determined using the elbow method, balancing the trade-off between cluster count and data compactness. Each cluster represents meaningful tissue patterns. Secondly,  we then compute Fisher vector encoding to model the distribution of feature embeddings within each cluster. Fisher vectors provide a robust representation of the distribution by modeling the feature space with a Gaussian mixture model (GMM). The Fisher vector for a set of descriptors \( \{ f(x_1), \dots, f(x_n) \} \) is formulated as:
\[
\frac{1}{n} \sum_{i=1}^{N} FV\left( f(x_i); v_1, \dots, v_m \right)
\]
where the Fisher Vector (FV) function is defined as:
\vspace{-5mm}

\begin{align*}
FV(.) = & \left[ 
\frac{s_{i1}}{c_1} \left( f(x_i) - v_1 \right), \dots, \frac{s_{im}}{c_m} \left( f(x_i) - v_m \right), \right. \\
& \left. \frac{s_{i1}}{\hat{c}_1} \left( f(x_i) - v_1 \right)^2,
\dots, \frac{s_{im}}{\hat{c}_m} \left( f(x_i) - v_m \right)^2 \right]
\end{align*}



In this equation, \(c_j\) and \(\hat{c}_j\) are constants used to scale the feature distributions. During GMM fitting, we set the number of coding centers to 5, ensuring a balance between model complexity and performance. The Fisher vector captures both the mean and variance of the components of feature distributions in each cluster, providing a compact and informative representation for the clustered patch embeddings.

Finally, the Fisher vectors from all clusters are concatenated to form a single high-dimensional vector. This concatenated vector effectively encapsulates both local and global information from the whole-slide image (WSI) by aggregating features across all patches and clusters. The resulting vector is then input into a permutation-invariant classifier for final classification.

We tested several classifier architectures, including multi-layer perceptron (MLP),  Swin Tiny \cite{liu2021swin}, Attention Multi-Instance Learning (AMIL)\cite{ilse2018attention}, and ConvNeXt \cite{liu2022convnet}, to evaluate their effectiveness for the classification task. These models leverage the rich patch-level and global representations encoded by the Fisher vectors, leading to robust and accurate classification of WSIs.

\section{Datasets, Experiments, and Results}
We experimented with various datasets, backbones, and comparative methods, as described below.

We worked with four datasets, focusing on classification and prediction tasks using H\&E-stained whole-slide images (WSIs):
\begin{itemize}
    \item \textbf{TCGA-BRCA}: We utilized data from the TCGA-BRCA cohort, which consists of 92 slides. Out of 92 slides, 36 are HER2-, and 56 are HER2+.
    \item \textbf{Warwick HER2}: This dataset provided 52 WSIs for training and 34 for testing for the Warwick University HER2 challenge, with tasks including binary classification of HER2+ vs. HER2- cases and HER2 score prediction.
    \item \textbf{TCGA-LUAD}: The dataset consisted of 159 slides, including 79 with EGFR mutations and 80 without. For this task, we performed binary classification between EGFR mutations vs. Non-EGFR mutations. 
    \item \textbf{CAMELYON17}: Comprising 500 WSIs from multiple centers, the dataset includes four classes: Negative, Isolated Tumor Cells (ITC), Macro-metastases, and Micro-metastases. We applied a binary classification task to distinguish between Metastasis Positive (including ITC, Macro, and Micro) and Negative classes. 
    
\end{itemize}

\begin{table*}[ht!]
\begin{center}
\begin{tabular}{|c|c|c|c|c|c|c|c|c|}
\hline
 Methods &Backbone Feature Extractor	  &Accuracy & AUC & Precision & Recall& F1-score	\\
\hline
DFVC~\cite{akbarnejad2021deep} & ResNet-50~\cite{jian2016deep} &	0.63 &  - &	0.75&	0.63& 0.61\\
\hline
AMIL~\cite{ilse2018attention} & ResNet50~\cite{jian2016deep} &	0.66 & 0.70	& 0.75 & 0.67 & 0.71\\
\hline
 Proposed method & EfficientNetV2-S~\cite{eff} &	\textbf{0.72}&	\textbf{0.71} & \textbf{0.77}&	\textbf{0.75}&\textbf{0.76}\\
\hline
\end{tabular}
\end{center}  
\caption{Results (\%) on the Warwick dataset for HER2 score. The best performance is marked as bold.}
\label{table1}
\end{table*}

\begin{table*}[ht!]
\begin{center}
\begin{tabular}{|c|c|c|c|c|c|c|c|c|}
\hline
 Methods &Backbone Feature Extractor	  &Accuracy & AUC & Precision & Recall& F1-score	\\
\hline
Anand, et al.~\cite{anand2020deep} & Neural Network  &	{0.75}&0.82	& {0.80}&	{0.75}&0.77\\
\hline
AMIL~\cite{ilse2018attention} & ResNet-50~\cite{jian2016deep} &	0.78 & 0.80	& 0.75 &0.81& 0.80\\
\hline
 Proposed Method & RegNetY-3.2GF~\cite{DBLP:journals/corr/abs-2003-13678} &	\textbf{0.80}&	\textbf{0.83} & \textbf{0.91}&	\textbf{0.90}& \textbf{0.90}\\
\hline
\end{tabular}
\end{center}  
\caption{ Results (\%) on the Warwick dataset for HER2+ vs HER2- classification. Best performance is in bold.}
\label{table2}
\end{table*}

\begin{table*}[ht!]
\begin{center}
\begin{tabular}{|c|c|c|c|c|c|c|c|c|}
\hline
 Methods &Backbone Feature Extractor	  &Accuracy & AUC & Precision & Recall& F1-score	\\
\hline

Anand, et al.~\cite{anand2020deep} & ResNet-50~\cite{jian2016deep}  &	0.73 & 0.76 & 0.70 & 0.87 & 0.78\\
\hline
AMIL~\cite{ilse2018attention} & ResNet-50~\cite{jian2016deep} &	0.76 & 0.64	& 0.67 & 0.75 & 0.71\\
\hline
Sekhar, et al.~\cite{sekhar2024her2} & MoCo-v2~\cite{chen2020improved}  &	0.82&\textbf{0.85} & 0.77 & \textbf{0.91} & 0.83 \\
\hline
 Proposed Method & MoCo-v2~\cite{chen2020improved} &	\textbf{0.86}&	0.83 & \textbf{0.88}&	0.86 &\textbf{0.87}\\
\hline
\end{tabular}
\end{center}  
\caption{ Results (\%) on the TCGA-BRCA dataset for HER2+ vs HER2- classification. Best performance is in bold.}
\label{table3}
\end{table*}

\begin{table*}[ht!]
\begin{center}
\begin{tabular}{|c|c|c|c|c|c|c|c|}
\hline
 Dataset (Method) &Backbone Feature Extractor	  &Accuracy & AUC & Precision & Recall& F1-score	\\
\hline \hline
 TCGA-LUAD (AMIL~\cite{ilse2018attention})& RegNetY-3.2GF~\cite{DBLP:journals/corr/abs-2003-13678} & 0.72&	0.72 &0.75 & 0.71&0.73\\
\hline
 TCGA-LUAD (Proposed)& RegNetY-3.2GF~\cite{DBLP:journals/corr/abs-2003-13678}  & \textbf{0.84}&	\textbf{0.79 }&\textbf{0.75 }& \textbf{0.80}&\textbf{0.77}\\
\hline \hline
 CAMELYON17 (AMIL~\cite{ilse2018attention})& EfficientNetV2-S~\cite{eff} & 0.75&	0.70 &0.73 &0.73 &0.73\\
\hline
CAMELYON17 (Proposed)& EfficientNetV2-S~\cite{eff}  & \textbf{0.77}&	\textbf{0.71} &\textbf{0.73} &\textbf{0.73} &\textbf{0.73}\\
\hline
\end{tabular}
\end{center}  
\caption{ Results (\%) on the different datasets for binary classification(Metastasis detection for CAMELYON17 and Mutation prediction for TCGA-LUAD) with our proposed method. Best performance is in bold.}
\label{table4}
\end{table*}


We employed various pre-trained encoders, including SimCLR~\cite{chen2020simple}, ResNet50~\cite{jian2016deep}, EfficientNet~\cite{eff}, RegNet~\cite{DBLP:journals/corr/abs-2003-13678}, ConvNeXT Tiny~\cite{liu2022convnet}, and Swin Tiny~\cite{liu2021swin} -- all trained on ImageNet~\cite{5206848} -- to extract features from image patches. These features were clustered using K-Means with the elbow method to determine the optimal cluster count, reducing feature space dimensionality. Ablation and validation studies were performed to compare various feature encoders, and the best-performing encoder was selected for each task, enhancing classification accuracy and robustness. For our experiments, we set $k = 10$ clusters across all datasets. Fisher vector encoding was applied with five $m=5$ centers (performed ablation study with m=10,15 and 3 but found best at m=5), with encoding parameters $\pi_{m} = 0.2$ and $\sigma_m = 0.1$ as in \cite{Arun2020}. These hyperparameters are decided based on performance for test dataset.

To augment the clustered features, we applied feature scaling (range: 0.9 -- 1), jittering (jitter level: 0.01), and mixup ($\alpha = 0.2$). These enhanced features were processed through an Attention Multi-Instance Learning (AMIL) block and optimized using AdamW with a learning rate of 0.001 and weight decay of 0.0001.

For the Warwick dataset, our model outperformed baselines in HER2 scoring into three classes (negative, equivocal, positive) and binary HER2+ vs HER2- classification (Tables~\ref{table1} and ~\ref{table2}). Similarly, in the TCGA-BRCA dataset, our approach achieved superior performance for HER2 status classification (Table~\ref{table3}), demonstrating its robustness. To test the robustness of our approach further, we extended the experiments to the TCGA-LUAD dataset for EGFR mutation prediction and the CAMELYON17 dataset for metastasis detection, with solid results in both tasks for the proposed model (Table ~\ref{table4}). Overall, our patch-based feature extraction, K-means clustering, and Fisher vector encoding approach outperformed baselines in HER2 classification tasks and adapted well to mutation and metastasis prediction, showcasing its broad applicability in digital pathology.

\section{Conclusion and Discussion}
\label{sec:page}
\vspace{-10pt}
In this work, we propose an efficient approach for whole slide image (WSI) classification by combining patch-based feature extraction, K-means clustering, and Fisher vector encoding. Our method captures both local and global tissue features, providing a compact representation of the entire WSI. By transforming patch clusters into Fisher vector representations, we reduce the dimensionality while preserving important tissue heterogeneity, making it well-suited for large-scale WSI classification. Our approach showed excellent performance across various datasets, including HER2 scoring (Warwick, TCGA-BRCA), mutation prediction (TCGA-LUAD), and metastasis detection (CAMELYON17). The consistency of results across different tasks highlights the generalizability of the method. While it holds promise for clinical deployment, future work should assess real-time performance and explore interpretability to build trust in clinical applications.

\bibliographystyle{unsrt}
\bibliography{paper}

\end{document}